%% file: main.tex
\documentclass[letterpaper, 10 pt, conference]{ieeeconf}  

\pdfoutput=1                                              
\IEEEoverridecommandlockouts                              

\usepackage{cite}
\usepackage{amsmath,amssymb,amsfonts}
\usepackage{algorithmic}
\usepackage{graphicx}
\usepackage{caption}
\usepackage{subcaption}
\usepackage{textcomp}
\usepackage{xcolor}
\usepackage{siunitx}
\usepackage{float}
\usepackage{amsmath,amssymb}
\usepackage{hyperref}
\hypersetup{colorlinks,allcolors=black}

\usepackage{tikz}
\usepackage{textcomp}
\usepackage{lipsum}

\newcommand\copyrighttext{%
  \footnotesize \textcopyright 2020 IEEE. Personal use of this material is permitted. Permission from IEEE must be obtained for all other uses, in any current or future media, including reprinting/republishing this material for advertising or promotional purposes, creating new collective works, for resale or redistribution to servers or lists, or reuse of any copyrighted component of this work in other works. DOI: 10.1109/IV47402.2020.9304601}
\newcommand\copyrightnotice{%
\begin{tikzpicture}[remember picture,overlay]
\node[anchor=south,yshift=10pt,xshift=10pt] at (current page.south) {\fbox{\parbox{\dimexpr\textwidth-\fboxsep-\fboxrule\relax}{\copyrighttext}}};
\end{tikzpicture}%
}

\graphicspath{ {images/} }

\def\BibTeX{{\rm B\kern-.05em{\sc i\kern-.025em b}\kern-.08em
    T\kern-.1667em\lower.7ex\hbox{E}\kern-.125emX}}

\title{\LARGE \bf 3D-DEEP: 3-Dimensional Deep-learning based on elevation patterns for road scene interpretation}

\author{A. Hern\'andez$^{1}$, S. Woo$^{2}$, H. Corrales$^{1}$, I. Parra$^{1}$, E. Kim$^{2}$, D. F. Llorca$^{1}$ and M. A. Sotelo$^{1}$
\thanks{$^{1}$ Computer Engineering Department, Polytechnic School, University of Alcal\'a, Madrid,  Spain. \{alvaro.hernandezsaz, hector.corrales, ignacio.parra, david.fernandezl, miguel.sotelo\}@uah.es}%
\thanks{$^{2}$ Electrical and Electronic Engineering,  Yonsei University, Seoul,  South Korea. \{etkim, wsh112\}@yonsei.ac.kr}%
}

\begin{document}

\maketitle

\copyrightnotice 

\thispagestyle{empty}
\pagestyle{empty}

\input{abstract.tex}

\input{Introduction.tex} 

\input{method.tex} 

\input{results.tex} 

\input{conclusions.tex} 

\input{Acknowledgements.tex} 

\bibliographystyle{IEEEtran}
\bibliography{references}
\end{document}

%% file: abstract.tex
\begin{abstract}
Road detection and segmentation is a crucial task in computer vision for safe autonomous driving. With this in mind a new net architecture (3D-DEEP) and its end-to-end training methodology for CNN-based semantic segmentation is described along this paper for. The method relies on disparity filtered and LiDAR projected images for three-dimensional information and image feature extraction through fully convolutional networks architectures. The developed models were trained and validated over Cityscapes dataset using just fine annotation examples with 19 different training classes, and over KITTI road dataset. 72.32\% mean intersection over union (mIoU) has been obtained for the 19 Cityscapes training classes using the validation images.  On the other hand, over KITTI dataset the model has achieved an F1 error value of 97.85\% in validation and  96.02\% using the test images.
\end{abstract}

%% file: Introduction.tex
\section{Introduction \& Related Work}

Semantic segmentation research, which attempts to assign semantic labels to each pixel within an image, is a fundamental task in computer vision. It can be widely applied to the fields of augmented reality devices, autonomous driving, diagnostic medicine or robotics. 

In the case of autonomous driving, semantic segmentation plays a very important role in identifying the environment in which the vehicle circulates. Every element of the environment where the vehicle is located must be taken into account, but its importance is different. One should not take the same care, for example, for a tree as for another vehicle or a pedestrian. For that reason, it is essential that within the data that are obtained from the sensors it is known to identify that it is each element and that part is occupying in the scene. This identification can be done through semantic segmentation. In some cases, the data are entirely segmented as in Cityscapes \cite{Cordts2016Cityscapes}. In others, such as KITTI \cite{Fritsch2013ITSC}, only a part of it is segmented, in this case, the road.  Although in the latter case there is only one class to segment, this is not unimportant as it is trivial to assume that the segmentation of the road in an autonomous vehicle's artificial vision system is vital, as this will be the area through which it can only circulate.

For object identification at the pixel level within scenes, in our case of traffic scenes, convolutional neural networks are usually used. The first models of this type of networks used as input data three-channel images (RGB) and they return a probability map of the size of the input image for each of the classes on which it is being trained \cite{long2015fully}.

Later on, other sources of input data, such as three-dimensional information, have been added in order to improve the results obtained.  This three-dimensional information can be obtained in different formats depending on the device where it comes. For example, disparity maps in stereoscopic vision systems or point clouds in LiDAR systems.

Within convolutional neural networks, BiSeNet \cite{yu2018bisenet} is a new architecture for real-time semantic segmentation that uses RGB images as input data. Through the extraction of context and spatial information of the input data has obtained excellent results in the dataset Cityscapes, CamVid and COCO-Stuff.

\begin{figure}[t]
    \centering
    \includegraphics[width=\linewidth]{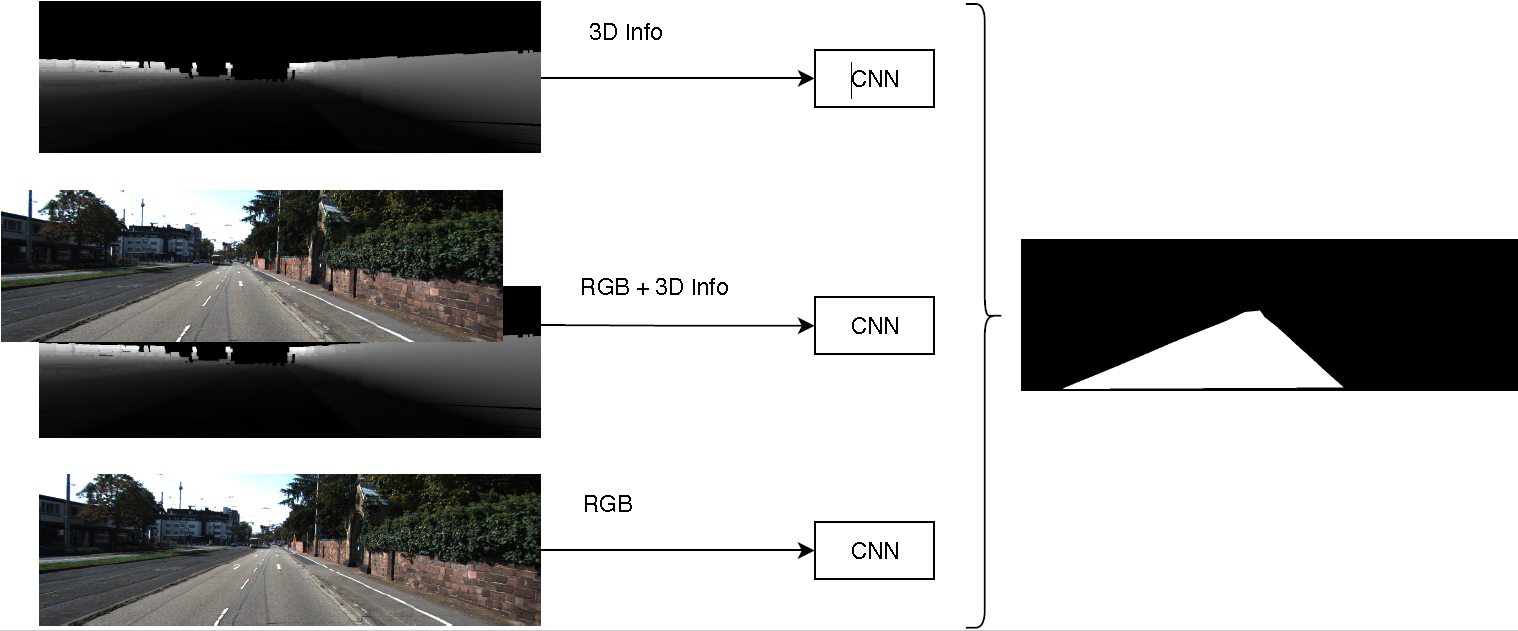}
    \caption{Flow diagram of the system.}
    \label{fig:Operation of the system}
\end{figure}

As can be seen in KITTI road benchmark \cite{Fritsch2013ITSC}, the first positions are occupied by methods of integrating three-dimensional information with RGB images. In this paper, a new network architecture is implemented that is capable of working with both three-dimensional information and two-dimensional information (Fig. \ref{fig:Operation of the system}). For this purpose, the BiSeNet architecture has been adapted, explained in more detail in the section \ref{Backbone}, so that it is capable of working with three-dimensional information sources along with RGB images with which it already worked in the beginning, keeping the processing in real-time and trying to improve its performance.

The rest of this document is organized as follows. The related papers are commented on the hereafter. The second section describes the methodology that led us to develop this new architecture, the different approaches to incorporate and treat the three-dimensional information and the function of each branch within the network.  In the third section, the line of training followed once the architecture has been developed, and the results obtained in the different sets of data used are presented. Finally, in the fourth section, the conclusions obtained from the results and possible future improvements are presented.

\subsection{Disparity maps}
Since their appearance along with stereo vision, disparity maps have been one of the most common ways to obtain three-dimensional information in a computer vision system. Their proven usefulness has allowed their use to be extended to multiple disciplines and tasks.
Besides, their lower cost compared to other devices makes these methods very competitive when determining their inclusion in a computer vision system. 

Within the field of autonomous vehicles, disparity maps are widely used in detecting damage to roads \cite{fan2019pothole}\cite{fan2019real}. In addition, they are also useful in detecting obstacles \cite{zhang2019outdoors}, drivable areas \cite{kim2019hw}.

Disparity maps are also widely used in semantic segmentation \cite{li2017traffic} where is used for segmentation in traffic scenes. In \cite{li2017study} they are also used to reduce noise at the outputs of a segmentation network and in \cite{palafox2019semanticdepth} they are generated from monocular vision and together with semantic segmentation they are used to estimate the position of the ego vehicle concerning the road.

\subsection{LiDAR Point clouds}

LiDAR devices rely on distance measurement through a set of laser beam arrays. Since one of its first non-military applications in meteorology \cite{goyer1963laser}, LiDAR devices have been in the cutting edge for 3D landscape mapping. These types of devices are not only used in autonomous driving projects, where they are better known; they have also produced outstanding results in other fields such as geology or biology.

Already in the field of autonomous driving, for example, in \cite{li2016vehicle} only point clouds obtained through LiDAR devices are used to detect the position of vehicles within the range of action through fully convolutional networks. In \cite{wu2019squeezesegv2}, a new network structure and point clouds are used for the segmentation of objects on the road and in \cite{behley2019semantickitti} where a dataset is created with KITTI point clouds and their respective labels for semantic scene understanding.

LiDAR devices are also widely used in the field of semantic segmentation. In \cite{aytaylan2018fully}, the fusion of hyperspectral images and point clouds is used for multiclass segmentation. By the other side, in \cite{wang2018pointseg} they use the data obtained directly from the LiDAR to form spherical images as input to a convolutional network toward its later segmentation. Finally, in \cite{pan2018semantic} they use a combination of high-resolution RGB images and LiDAR data for semantic segmentation of aerial landscapes.

This kind of devices can provide a great amount of information. Its combination with the features obtained from the RGB images has surpassed the previous methods being the first two positions occupied by RGB plus LiDAR methods (RGB-L), as shown in KITTI's road benchmark.

%% file: method.tex
\section{Method}

This section is focused on 3D-DEEP, the new fully convolutional network architecture proposed in this paper as a solution for semantic segmentation from RGB and 3D data inputs.

\subsection{Backbone}
\label{Backbone}

The architecture of 3D-DEEP net is an evolution of the previously mentioned BiSeNet \cite{yu2018bisenet}. BiSeNet focuses on harnessing spatial information and receptive field. Both approaches are critically important to achieve high accuracy results \cite{zhao2017pyramid}\cite{chen2017rethinking}. As a solution to solve both approaches, BiSeNet architecture proposes using two different branches. These branches are focused on acquiring all available information from the RGB image in each its respective field, spatial and context information. The branch in charge of maintaining enough resolution of the image to obtain adequate spatial information is the spatial path. On the other hand, the branch in charge of achieving the broadest receptive field to obtain sufficiently rich contextual information is the context path.

Although this approach is advantageous, it has been considered the possibility to improve it by incorporating another data source, such as three-dimensional information synchronized with images. The idea of using raw point clouds together with their respective images as input was discarded because it would substantially increase the complexity of the problem and its computational costs since it would need to use three-dimensional convolutional layers. Besides, three-dimensional information can come from different sensors, and our idea was that our architecture should not be limited to one, allowing the network to be as flexible as possible.

Therefore, having dismissed the use of merely projecting the point cloud in the two-dimensional plane, it became necessary to find a way of using the data to obtain as much information as possible and not making the network incompatible with other sources of three-dimensional information such as disparity maps.

There are many possibilities to work with a point cloud projection on the image plane. Projecting it and make its intensity or chromatic range coincide with the absolute distance of the projected point to the system's coordinate origin is the most trivial. Although it can be useful in many cases, it has been considered that this type of projection does not provide the appropriate information given the task to be performed with the KITTI data set from which the point clouds have been obtained. The task for the selected KITTI dataset is the semantic segmentation of the road, and the distance of each point of the road to the origin of the coordinates will usually be a very variable value.

\begin{figure}[t]
\vspace{0.3cm}
     \centering
     \begin{subfigure}{\linewidth}
         \centering
         \includegraphics[width=\textwidth]{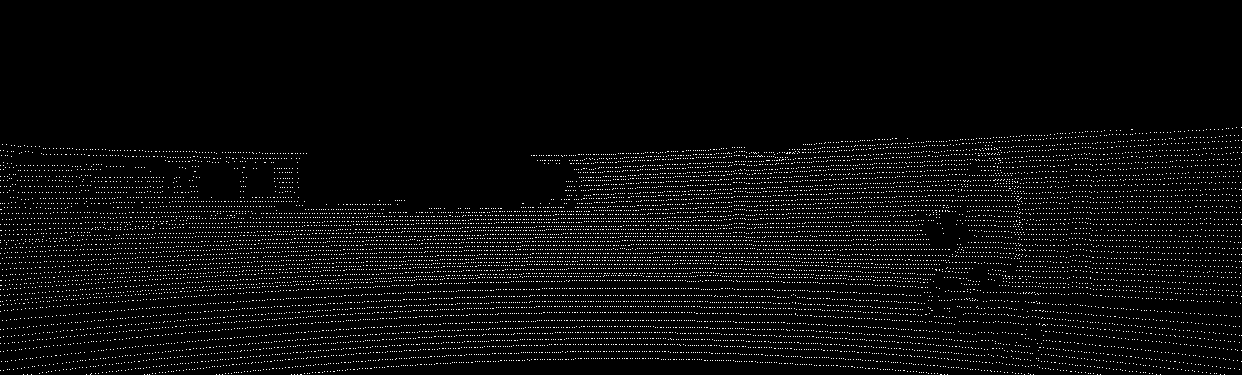}
         \caption{Normalized LiDAR projected points}
         \label{fig:pointdiff}
     \end{subfigure}
     \hfill
     \begin{subfigure}{\linewidth}
         \centering
         \includegraphics[width=\textwidth]{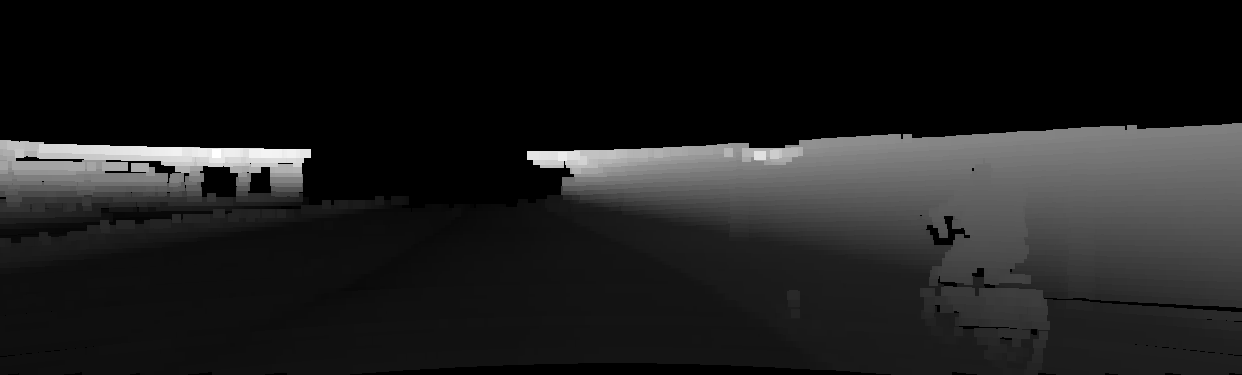}
         \caption{Elevation pattern image}
     \end{subfigure}
        \caption{Above, LiDAR projected points according to their elevation. Below, after applying the dilation.}
        \label{fig:altimgs}
\vspace{-0.3cm}
\end{figure}

In \cite{chen2019progressive} they observed that the LiDAR points belonging to the road usually have a shallow height concerning the reference axis situated on the LiDAR device and do not have substantial changes of height. Then, from the projection of the points and their height they develop an algorithm that calculates the correlation between the height of the neighbouring pixels, which provides images that can be interpreted as mean absolute value for gradients of elevations of projected points.

With the idea of elevation in mind, a direct method to create elevation difference patterns through LiDAR points is proposed. As in \cite{chen2019progressive}, the raw LiDAR points are assumed and using the calibration parameters they are projected in the 2D image plane. In turn, the intensity of each projected point is the normalized value from 0 to 255 of the LiDAR Z coordinate value, eq(\ref{eq:1}).
To deal with Z outliers, only those LiDAR points that are within a range of coordinates and a field of view are taken into account. The horizontal fov is taken from $-60^{\circ}$ to $60^{\circ}$ and the vertical from $-13.9^{\circ}$ to $2.9^{\circ}$. The ranges of coordinates for x, y, z values are $[0, 80], [-60, 60], [-2.1, 2.9]$ respectively. This values has been empirically obtained. As an additional measure, points that once projected on the image plane are outside its dimensions are also excluded.

\begin{equation} \label{eq:1}
   Zn = \frac{Z - Z min}{Z max -  Z min} * 255
\end{equation}

However, even though at first glance, the projected image reflected gives enough information to deduce where the road is, the distance between laser beams creates too many black spaces that not provide any information. To solve the lack of information a dilation operation is applied to the image with a 9 x 9 size kernel that satisfactorily completes practically all the gaps left by the resolution of the LiDAR without significant impact in non zero values of the image. This transformation gives us what we have called an elevation pattern image (elvdiff), a grayscale image in which the elevation differences obtained from the LiDAR data are adequately represented \ref{fig:altimgs}. In Fig. \ref{fig:pointdiff} the intensity of the projected points was modified for visibility reasons.

\subsection{Branches}

\begin{figure}[t]
\vspace{0.3cm}
    \centering
    \includegraphics[width=\linewidth]{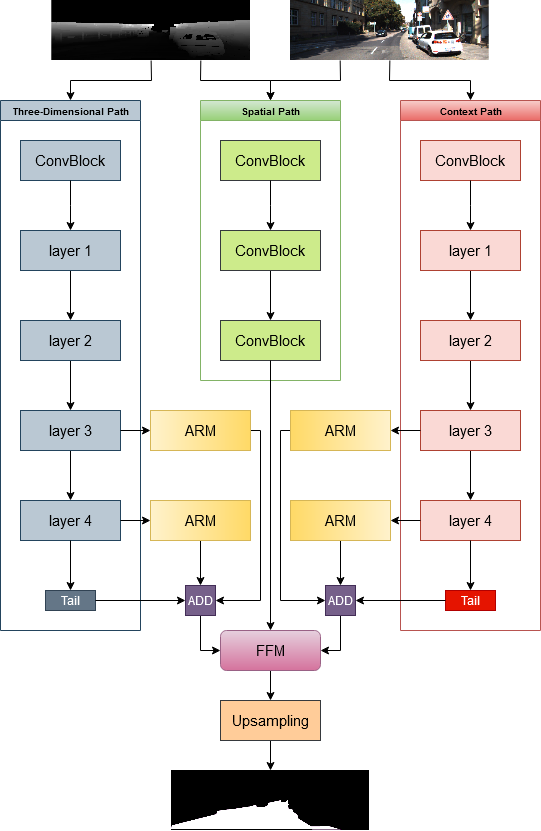}
    \caption{3D-DEEP network architecture.}
    \label{fig:Net architecture}
\vspace{-0.3cm}
\end{figure}

Once the LiDAR data have been adapted, the next step was to modify the architecture of the initial network proposed in BiSeNet to work not only with RGB images but also with three-dimensional information through disparity maps or the previously proposed elvdiff images. The idea that emerged for this was that just as \cite{yu2018bisenet} uses a different branch for each type of information it wants to collect, the 3D-DEEP architecture could do the same. Therefore, a new architecture with three branches was implemented, adding to the previous structure the three-dimensional branch, where most of the information coming from this source is processed, as shown in Fig. \ref{fig:Net architecture}.

\subsubsection{Context Path}

The Context Path is designed to provide sufficient receptive field to achieve optimal performance levels while still taking into account computational efficiency. For this purpose, the ResNet architectures proposed in \cite{he2016deep} are considered as a context path. This choice was made due to several factors: the ease of integrating this network within the complete context path structure, the excellent performance it has had and the possibility of easily comparing the results with different levels of network depth.

Since this branch is in charge of extracting the context information of the image, the input data is the RGB values, so the number of the input channels in the first convolutional layer should be three. Then the image pass through each one of the five convolutional layers of the architecture until obtaining a features vector of 2048 of depth in the architectures 50, 101 and 152 and of 512 in the architectures 18 and 34, on which a global average pooling is computed to obtain the tail features vector.
Besides, the feature vectors obtained in layer three and layer four are sent to the attention refinement module for their subsequent incorporation into the feature fusion module with the previously mentioned tail.

\subsubsection{Three-dimensional Path}

This branch is the twin of the context path and, like its counterpart, is in charge of obtaining a large receptive field from the three-dimensional data supplied. Since its function is quite similar to the context path, the backbone net is still the ResNet architectures. As mentioned above, it has been considered that the branch in charge of three-dimensional information processing should not be completely limited by the data input format. Achieving complete independence is impossible, but we believe that a network capable of processing images in which three-dimensional information has been dumped both in a particular way as in the elvdiff images of our case and in a more general way through disparity maps, was very close to the idea of flexibility concerning the input of three-dimensional data.

Unlike the context path, the three-dimensional path data is provided in single-channel grayscale images, so the first convolutional layer of the ResNet architecture has been modified to accept them.
Since the width and height of the images that store the three-dimensional data are the same as that of the RGB images that are analyzed in the context path, the depth of the feature vectors of both architectures 50, 101 and 152, and architectures 18 and 34 will be the same, so their incorporation to the attention modules and feature fusion module is equivalent to the one used in the context path. That particularity allows us to reuse the same modules without any extensive amendments (see Fig. \ref{fig:Net architecture}).  

\subsubsection{Spatial path}

The spatial path is the central branch that appears in Fig. \ref{fig:Net architecture} and its in charge of providing enough spatial information from the data. In semantic segmentation, exist specific approaches for providing sufficient image resolution to enable enough spatial information to be encoded through the dilated convolution as in \cite{zhao2017pyramid}\cite{yu2015multi}. However, in order to try to maintain the processing speed achieved by BiSeNet and not increase memory comsumption, it has been chosen to use the same architecture used in \cite{yu2018bisenet}.

As in BiSeNet architecture, the spatial path contains three layers. However, these contain slight differences that make them suitable for input data.  Each layer includes a convolution with stride 2, followed by batch normalization \cite{ioffe2015batch} and ReLU \cite{glorot2011deep}. In order to adapt the layers to the RGB-L and RGB-D data inputs of the new network, the number of input channels of the first convolutional layer was increased from three to four, as was the initial idea of the architecture. After checking that the results were favourable, the idea of duplicating the spatial path was discarded in order to maintain the processing speed. Also, there is the option to set the padding value in the first layer depending on the chosen input data. If the images belong to Cityscapes (2048x1024) or KITTI in bird's eye view (400x800) the padding will be two, being one for KITTI images in perspective (1242x375).

These modifications allow maintaining the initial idea of making prevail the original spatial size of the data and coding sufficient spatial information since the size of the output feature maps of 1/8 of the original size of the image is maintained. In our opinion, these maps are large enough to obtain rich spatial information.

\subsubsection{Feature fusion module and attention refinement module}

The feature fusion and attention refinement modules maintain the same functionality as in \cite{yu2018bisenet} as shown in Fig. \ref{fig:Net architecture}. The characteristics obtained from the spatial path, the context path and the three-dimensional path are situated in different representation levels so, as indicated in BiSeNet, they cannot be added directly and there lies the utility of the feature fusion module. In turn, the attention refinement module is in charge, as in BiSeNet, of refining the characteristics of the last two convolution phases of the context path integrating the global context information easily without any up-sampling operation.

%% file: results.tex
\section{Training \& Results}
This section will discuss the training methods that have been used for the 3D-DEEP network and their corresponding results in the KITTI \cite{Fritsch2013ITSC} and Cityscapes \cite{Cordts2016Cityscapes} dataset.

For the Cityscapes training, 2975 images from the left camera were used along with their respective disparity maps. These images belong to the set of images whose ground truth is finely annotated from the semantic segmentation benchmark. With the idea of improving the disparity maps, the technique proposed in \cite{min2014fast} is applied to try to complete the information available in the original maps.

For the training with KITTI, the 289 road images provided in their semantic segmentation road dataset and their respective point clouds were used. Since no validation images are provided, Monte Carlo cross validation \cite{xu2001monte} was used during the training as a method of analyzing the results. Thirty images with their respective point clouds are randomly separated in each iteration and used as a validation set. The number of iterations considered for each training is four. This number will be increased to 10, with the idea of generating much more stable results, once all the hyper-parameters that previously gave the best results have been decided. The training was done in the graphic card  Nvidia Titan RTX.

In both datasets, Kaiming initialization is used for those layers that, like spatial path, do not use Imagenet's pre-trained weights. The number of trainable parameters pending of the context path model is shown in table \ref{n_parameters}.

\begin{table}[b]
    \caption{Number of trainable parameters according to the backbone model}
    \begin{center}
        \begin{tabular}{c|c}
            \textbf{Backbone} & \textbf{Number of trainable parameters} \\ \hline
            ResNet 18 & \num{1,31E+08} \\
            ResNet 34 & \num{1,00E+08} \\
            ResNet 50 & \num{6,21E+07} \\
            ResNet 101 & \num{4,47E+07} \\
            ResNet 152 & \num{2,44E+07}
        \end{tabular}
        \label{n_parameters}
    \end{center}
\end{table}

\begin{figure}[t]
    \centering
    \includegraphics[width=\linewidth]{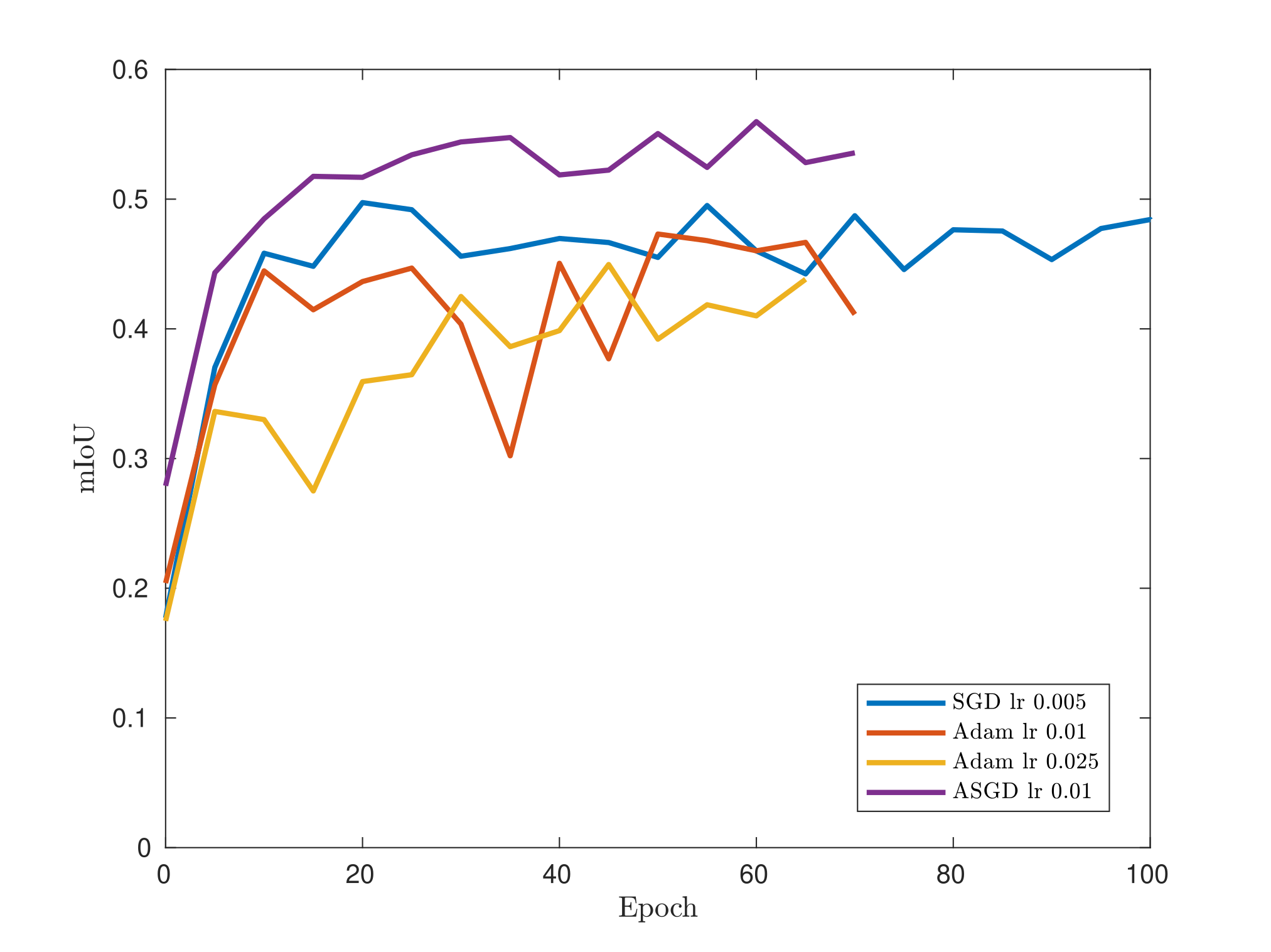}
    \caption{Evolution of mIoU on training for different optimizers.}
    \label{fig:SGD Training}
\end{figure}

\vspace{-0.75cm}

\subsection{Cityscapes}

As a starting point, our first training was done with the ResNet18 network as the backbone for the context path, using a stochastic gradient descent optimizer and with \num{5e-3} as the learning rate. The results, as can be seen in Fig. \ref{fig:SGD Training} were not satisfactory as from epoch 20 the mIoU does not improve, and its maximum value is close to 50\%. The change of optimization method to Adam did not get the desired results either, as it can be seen in the same graph, even varying the learning rate slightly. In both runs, the maximum values remained around 45\%. The optimization algorithm that worked best in this first attempt was the ASGD \cite{polyak1992acceleration}, which showed a more stable trend and reached maximum values for mIoU of 56\%. These are still not very good but have been considered an acceptable starting point, so the following pieces of training were done with ASGD.

\begin{figure}[t]
    \centering
    \includegraphics[width=\linewidth]{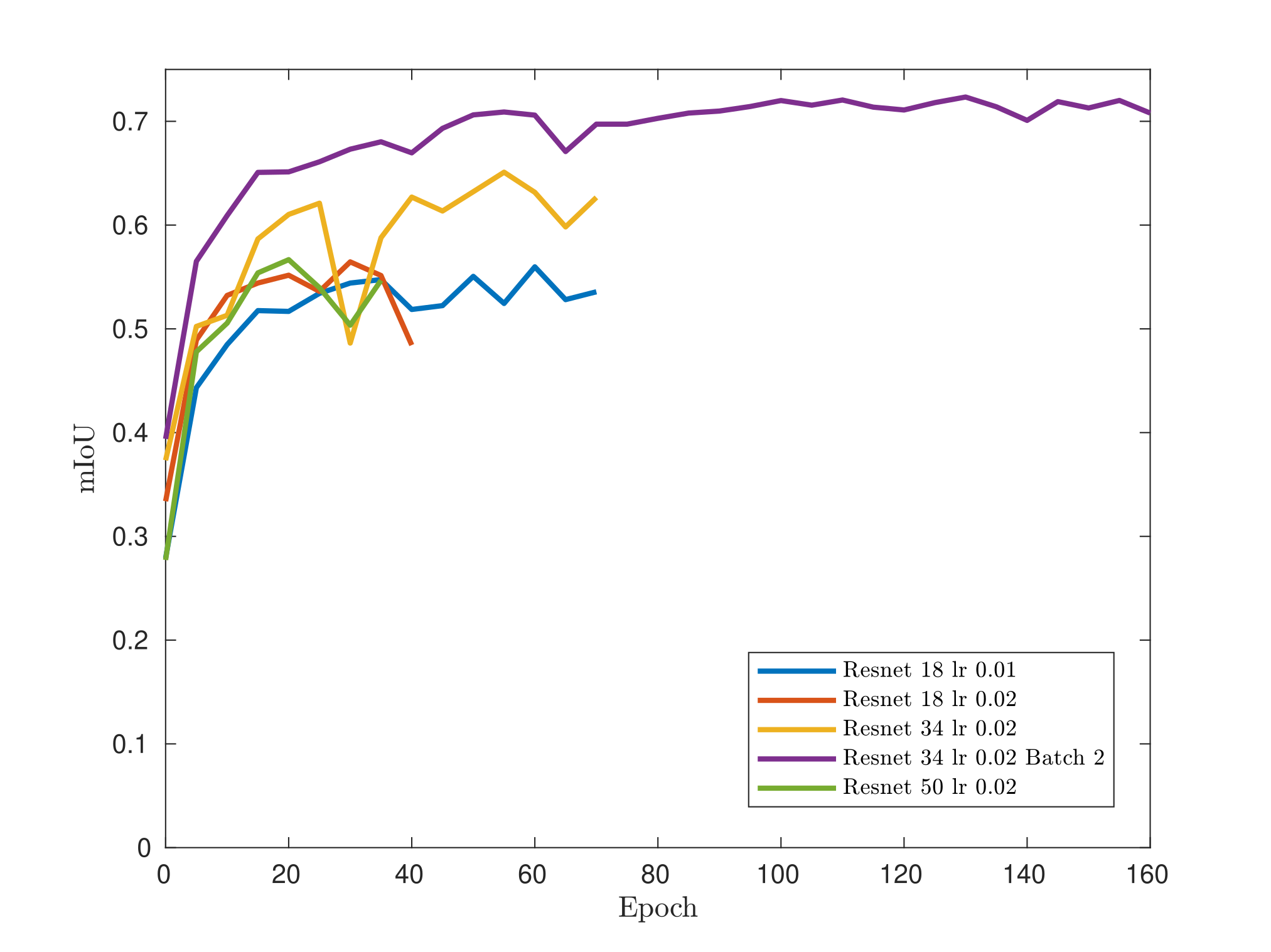}
    \caption{Evolution of mIoU on training for different learning rates and context paths.}
    \label{fig:Backbone Training}
\end{figure}

Once the optimization algorithm was decided, the following training sessions were oriented towards choosing the optimal hyperparameters and backbone for the context and three-dimensional paths. Due to memory limitations, a heavy learning strategy was implemented\cite{7478072}. As shown in Fig. \ref{fig:Backbone Training}, the learning rate remained close to the initial values, but it was decided to slightly increase it since the three-dimensional branch did not have pre-trained weights. Besides, from this moment on, data augmentation was used to look for a higher generalization and therefore, better validation results. For data augmentation it was used mirroring, salt and pepper noise, Poisson noise, speckle noise, blurring, colour casting, colour jittering, affine transformations, perspective transformations, cropping, shear and rotations.

The results obtained indicated that the optimum learning rate was around 0.02 and that the ideal backbone was ResNet34. Increasing the depth, with ResNet50, worsened the results probably because the images had to be resized.

Two improvements were applied to the initial version. First, using pre-trained weights in imagenet for the three-dimensional branch. Second, using the APEX \cite{Apex2019} library to be able to establish a minibatch of two images. These increased the  mIoU to 72.34\%, as depicted in Fig. \ref{fig:Backbone Training}.

Although the results were acceptable (they would have ranked second in Cityscapes among the algorithms using the same data types) they were not considered good enough to be ranked in Cityscapes.  The results in test is usually worse, so more work was put on the system until a way to improve the results was found.

\subsection{KITTI}

The training with KITTI images was less restrictive in memory consumption thanks to the smaller size of its images, so it was possible to test all ResNet architectures and increase the size of the minibatch to four. Furthermore, as the training was after the Cityscapes one, everything learned previously was adapted so that the process was much more efficient in the search for optimal results. Therefore, from now on,  training is considered to be done with ASGD and all the branches with pre-trained weights.

The first training with images belonging to KITTI's dataset, which was taken as a starting point, already obtained excellent results, being the minimum F1 value of 96.16\% for the ResNet18 architecture and the maximum F1 value 96.80\% for the ResNet152 architecture. 

As an idea to improve the obtained results for the moment, it was determined to use a strategy of learning rate schedule. For this purpose, it was decided to implement the model proposed in \cite{smith2017cyclical} of cyclical and decreasing cyclical variation of the learning rate with triangular functions. The training was performed with an upper limit value of 0.25 and a lower limit value of 0.0001 previously calculated. However, this did not lead any results improvement, in some cases, even worsening on the order of one percentage point. Nevertheless, the limits of the learning rate obtained in the course of these tests provided us with useful information for the next steps.
Continuing with the idea of establishing a scheduling policy, it was decided to use the upper limit and its three immediate divisions by two as the initial value (0.125, 0.0625, 0.03125) for the learning rate with a decreasing polynomial scheduler, so that it would come close to the previously obtained lower limit. This idea, as can be seen in Table \ref{Training results}, substantially improved the results, enough to consider ranking them in the KITTI benchmark.

\begin{table}[t]
\vspace{0.2cm}
    \caption{Training results in the KITTI dataset with polynomial scheduler for the learning rate}
    \begin{center}
        \begin{tabular}{c|ccc}
        \textbf{Backbone} & \textbf{Initial lr} & \textbf{F1 value} & \textbf{Max F1} \\ \hline
        ResNet152 & 0.25 & 96.87\% & 97.3 \\
        ResNet50 & 0.125 & \textbf{97.37\%} & 97.66 \\
        ResNet34 & 0.0625 & 97.33\% & \textbf{97.77} \\
        ResNet152 & 0.03125 & 97.11\% & 97.35
        \end{tabular}
    \label{Training results}
    \end{center}
\vspace{-0.6cm}
\end{table}

Ranking in KITTI posed a problem, implied to transform the images to a bird's eye view and all the training had been done in a panoramic view. Initially, the probability map returned by the network was transformed to a bird's eye view, but the results were not good. Therefore it was decided to transform the whole dataset to BEV (using the code provided with the dataset) and train directly from that perspective, obtaining outstanding results, with maximum values of around 98\% for the F1 value. 

As the last step, ten training sessions were carried out with the best configuration obtained for the moment, also using Monte Carlo Cross Validaton \cite{xu2001monte}, with the idea of obtaining an average value that would be as close as possible to the one obtained with the test images. This training gave an average value for the F1 error of 97.09\% and maximum values close to 98\% between all the iterations.
\vspace{-0.1cm}
\begin{table}[htp]
    \caption{KITTI dataset Evaluation Results in test images (percentage)}
    \begin{center}
        \begin{tabular}{c|cccccc}
        \textbf{Benchmark} & \textbf{MaxF} & \textbf{AP} & \textbf{PRE} & \textbf{REC} & \textbf{FPR} & \textbf{FNR} \\ \hline
        UM & 95.35 & 93.50 & 95.20 & 95.51 & 2.20 & 4.49 \\
        UMM & 97.27 & 95.76 & 97.01 & 97.54 & 3.31 & 2.46 \\
        UU & 94.67 & 93.04 & 94.23 & 95.12 & 1.90 & 4.88 \\
        URBAN & 96.02 & 94.00 & 95.68 & 96.35 & 2.39 & 3.65
        \end{tabular}
    \label{kittiresults}
    \end{center}
\end{table}
\vspace{-0.5cm}

Being the optimal configuration ResNet101 as backbone with a learning rate of 0.03125, the results were ranked obtaining the eighth position tied with the seventh as shown in Table \ref{kittiresults}. Going into more detail, Table \ref{UMM_results} shows that our network is twenty hundredths above second in AP within the UMM\_ROAD category. This category, in which the network proposed in this article has its best results, includes urban roads with multiple marked lanes. For further reference see \href{http://www.cvlibs.net/datasets/kitti/eval_road_detail.php?result=31f1ca0b557accf872536151597816e957af9c51}{3D-DEEP KITTI results}.

\vspace{-0.1cm}

\begin{table}[hbp!]
    \caption{Comparison of KITTI evaluation results in UMM (percentage)}
    \begin{center}
        \begin{tabular}{c|cccc}
        \textbf{Method} & \textbf{MaxF} & \textbf{AP} & \textbf{PRE} & \textbf{REC} \\ \hline
        3D-DEEP (Ours) & 97.27 & \textbf{95.76} & 97.01 & 97.54 \\
        PILARD\cite{chen2019progressive} & 97.77 & 95.64 & 97.75 & 97.79 \\
        LidCamNet\cite{caltagirone2019lidar} & 97.08 & 95.51 & 97.28 & 96.88
        \end{tabular}%
    \label{UMM_results}
    \end{center}
\end{table}

\begin{table}[t]
\vspace{0.2cm}
    \caption{Comparison of KITTI evaluation results in Urban (UM+UMM+UU). (percentage)}
    \begin{center}
        \begin{tabular}{c|cccc}
        \textbf{Method} & \textbf{MaxF} & \textbf{AP} & \textbf{PRE} & \textbf{REC} \\ \hline
        PILARD\cite{chen2019progressive} & 97.03 & \textbf{94.03} & 97.19 & 96.88 \\
        3D-DEEP (Ours) & 96.02 & 94.00 & 95.68 & 96.35 \\
        LidCamNet\cite{caltagirone2019lidar} & 96.03 & 93.93 & 96.23 & 95.83
        \end{tabular}%
    \label{URBAN_results}
    \end{center}
\vspace{-0.2cm}
\end{table}

\begin{table}[t]
    \caption{Comparison of evaluation results in KITTI for the fastest architectures (percentage)}
    \begin{center}
        \begin{tabular}{c|ccccc}
        \textbf{Method} & \textbf{MaxF} & \textbf{AP} & \textbf{PRE} & \textbf{REC} & \textbf{Runtime} \\ \hline
        NF2CNN & \textbf{96.70 } & 89.93  & 95.37  & 98.07  & 0.006 s \\
        ChipNet\cite{lyu2018chipnet} & 94.05  & 88.29  & 93.57  & 94.53  & 0.012 s \\
        LoDNN\cite{caltagirone2017fast} & 94.07  & 92.03  & 92.81  & 95.37  & 0.018 s \\
        multi-task CNN\cite{oeljeklaus2018fast} & 86.81  & 82.15  & 78.26  & 97.47  & 0.0251 s \\
        ALO-AVG-MM\cite{reis2019combining} & 92.03  & 85.64  & 90.65  & 93.45  & 0.0296 s \\
        LCFNet & 96.42  & 91.05  & 96.60  & 96.24 & 0.03 s \\
        3D-DEEP (Ours) & 96.02  & \textbf{94.00} & 95.68 & 96.35 & 0.03 s
        \end{tabular}%
    \label{fastest_results}
    \end{center}
\vspace{-0.6cm}
\end{table}
\vspace{-0.5cm}

Within the general category URBAN\_ROAD, which includes the categories UU (Urban Unmarked), UM (Urban Marked), UMM (Urban Multiple Marked), the network is only four-hundredths of the first position in the average accuracy, it has been considered an excellent result. (see table \ref{URBAN_results})
Considering the fastest methods officially classified, 3D-DEEP ranks first in AP and third in MaxF, with more than acceptable results for implementation in a real-time system. (see table \ref{fastest_results}).
\vspace{-0.1cm}

\begin{figure}[htp!]
    \centering
    \begin{subfigure}{\linewidth}
        \centering
        \includegraphics[width=\textwidth]{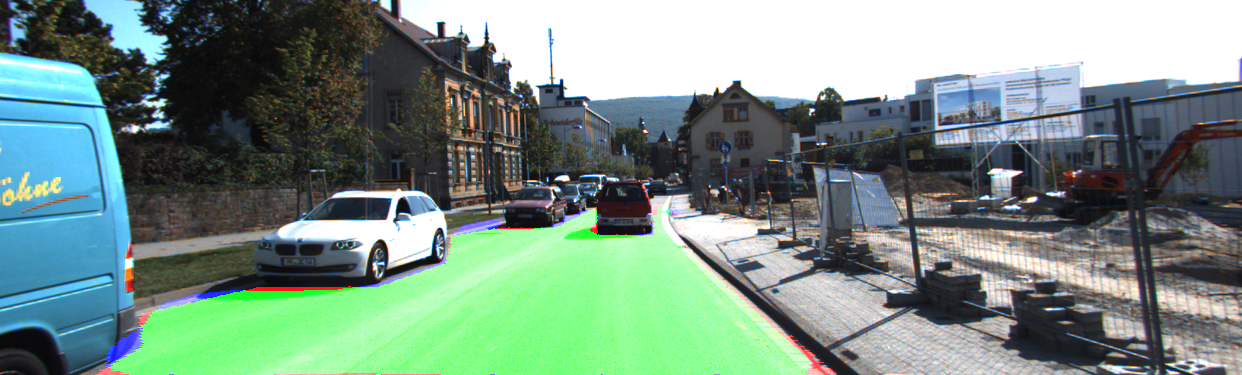}
        \caption{Urban unmarked road}
    \end{subfigure}
    \hfill
    \begin{subfigure}{\linewidth}
        \centering
        \includegraphics[width=\textwidth]{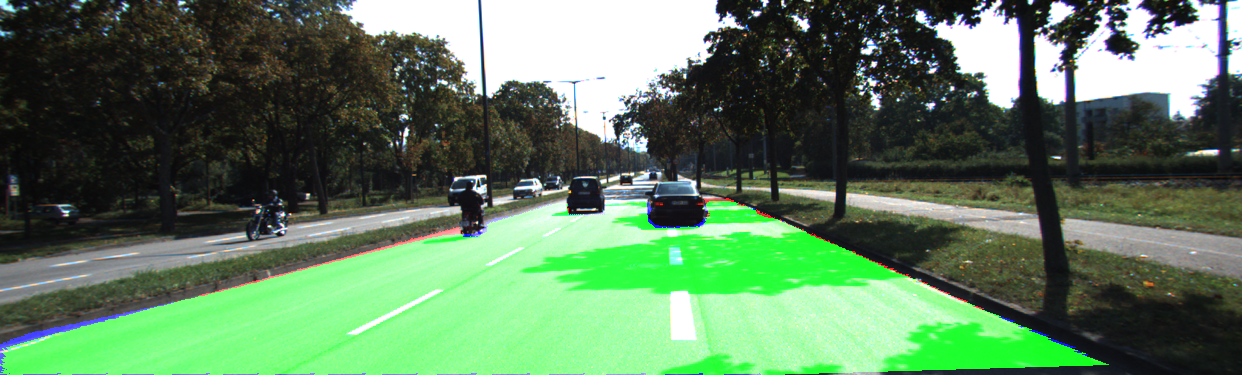}
        \caption{Urban multiple marked road}
    \end{subfigure}
    \hfill
    \begin{subfigure}{\linewidth}
        \centering
        \includegraphics[width=\textwidth]{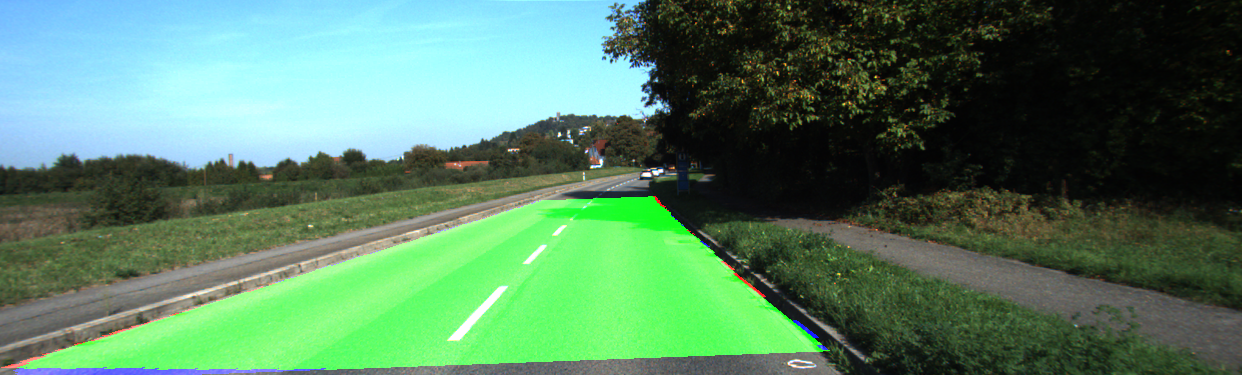}
        \caption{Urban marked road}
     \end{subfigure}
        \caption{Results for KITTI dataset in test images.}
        \label{fig:Result images}
\end{figure}
\vspace{-0.1cm}

Quantitatively the results are also excellent for each category evaluated, as shown in Fig. \ref{fig:Result images}. The network detects the majority of the labelled road in the test images (pixels labelled green) with hardly any false positives (pixels labelled blue) and false negatives (pixels labelled red). These qualitative results, together with the corresponding quantitative ones, allow 3D-DEEP to consider an optimal architecture for implementation in an autonomous driving system.

%% file: conclusions.tex
\section{Conclusions}
Several conclusions can be drawn from this work. We are satisfied with the performance of the network proposed in this document, especially when working with data obtained through point clouds. When three-dimensional data is obtained from the stereoscopic vision, the performance is not good enough, so more research should be done in that field to make 3D-DEEP more versatile. Another possible improvement point is to adapt the network so that in the context path newer architectures like ResNext or Inception can be used, which could further improve the already remarkable results obtained with LiDAR and make the results obtained with disparity maps more in line with state of the art.

%% file: Acknowledgements.tex
\section{Acknowledgements}
This work was supported in part by Spanish Ministry of Science, Innovation and Universities (Research Grant DPI2017-90035-R) and in part by the Community Region of Madrid (Research Grant 2018/EMT-4362 SEGVAUTO 4.0-CM) and BRAVE Project, H2020, Contract \#723021. This project has received funding from the Electronic Component Systems for European Leadership Joint Undertaking under grant agreement No 737469 (AutoDrive  Project) and in part by the Spanish Ministry of Economy (Research Grant PCIN-2017-086). This Joint Undertaking receives support from the European Unions Horizon 2020 research and innovation programme and Germany, Austria, Spain, Italy, Latvia, Belgium, Netherlands, Sweden, Finland, Lithuania, Czech Republic, Romania, Norway. We gratefully acknowledge the support of NVIDIA Corporation with the donation of the GPUs used for this Research.